\documentclass{article}
\usepackage{amsmath,graphicx,mlspconf}
\usepackage{xcolor}
\usepackage{tabularx,txfonts}

\usepackage{hyperref,url,booktabs,amsfonts,nicefrac,microtype,xcolor,tabularx,txfonts,bm,appendix,mathtools,algorithm,algorithmic,subfig,array,wrapfig,enumitem,colortbl,inputenc,fontenc,caption}

\usepackage{xcolor}
\definecolor{grey}{rgb}{0.5, 0.5, 0.5}

\usepackage{microtype}
\usepackage{graphicx}
\usepackage{subfig}
\usepackage{amsmath}
\usepackage{amssymb}
\usepackage{cancel}
\usepackage{xcolor}
\usepackage{multirow}
\usepackage{comment}
\usepackage{booktabs}
\usepackage{etoolbox}

\copyrightnotice{979-8-3503-2411-2/25/\$31.00 {\copyright}2025 IEEE}

\toappear{2025 IEEE International Workshop on Machine Learning for Signal Processing, Aug.\ 31-- Sep.\ 3, 2025, Istanbul, Turkey}

\title{SVD Based Least Squares for X-Ray Pneumonia \\ Classification Using Deep Features}

\name{{Mete Erdogan}, {Sebnem Demirtas}}
\address{
College of Engineering, Koc University, Istanbul, Turkey \\
\textit{\normalsize{\{merdogan18, sdemirtas20\}@ku.edu.tr}}
}

\begin{document}
\maketitle

\begin{abstract}
Accurate and early diagnosis of pneumonia through X-ray imaging is essential for effective treatment and improved patient outcomes. Recent advancements in machine learning have enabled automated diagnostic tools that assist radiologists in making more reliable and efficient decisions. In this work, we propose a Singular Value Decomposition-based Least Squares (SVD-LS) framework for multi-class pneumonia classification, leveraging powerful feature representations from state-of-the-art self-supervised and transfer learning models. Rather than relying on computationally expensive gradient-based fine-tuning, we employ a closed-form, non-iterative classification approach that ensures efficiency without compromising accuracy. Experimental results demonstrate that SVD-LS achieves competitive performance while offering significantly reduced computational costs, making it a viable alternative for real-time medical imaging applications. The implementation is available at: \href{https://github.com/meterdogan07/SVD-LS}{\text{github.com/meterdogan07/SVD-LS}}.
\end{abstract}


\section{Introduction}

Pneumonia is a critical respiratory condition and one of the foremost causes of morbidity and mortality worldwide, especially among children under five and the elderly \cite{who_pneumonia}. Prompt and accurate diagnosis is vital to ensure effective treatment and reduce the risk of complications. Chest X-ray imaging remains a standard and cost-effective diagnostic tool in clinical practice. However, the manual interpretation of chest X-ray images is time-intensive and subject to inter-observer variability, motivating the development of automated tools that can assist radiologists in clinical decision-making. \\

Recent advances in computer vision have led to the widespread adoption of deep learning models such as Convolutional Neural Networks (CNNs) \cite{lecun1989backpropagation, resnet} and Vision Transformers (ViTs) \cite{dosovitskiy2020image} for pneumonia detection from X-ray images. These models achieve impressive performance but often rely on large-scale annotated datasets, intensive training, and/or fine-tuning procedures. Such computational and data demands may limit their deployment in resource-constrained healthcare settings or real-time applications. \\

\begin{figure}[t!]
    \centering
    \includegraphics[trim = {0cm 0cm 0cm 0cm},clip,width=0.45\textwidth]{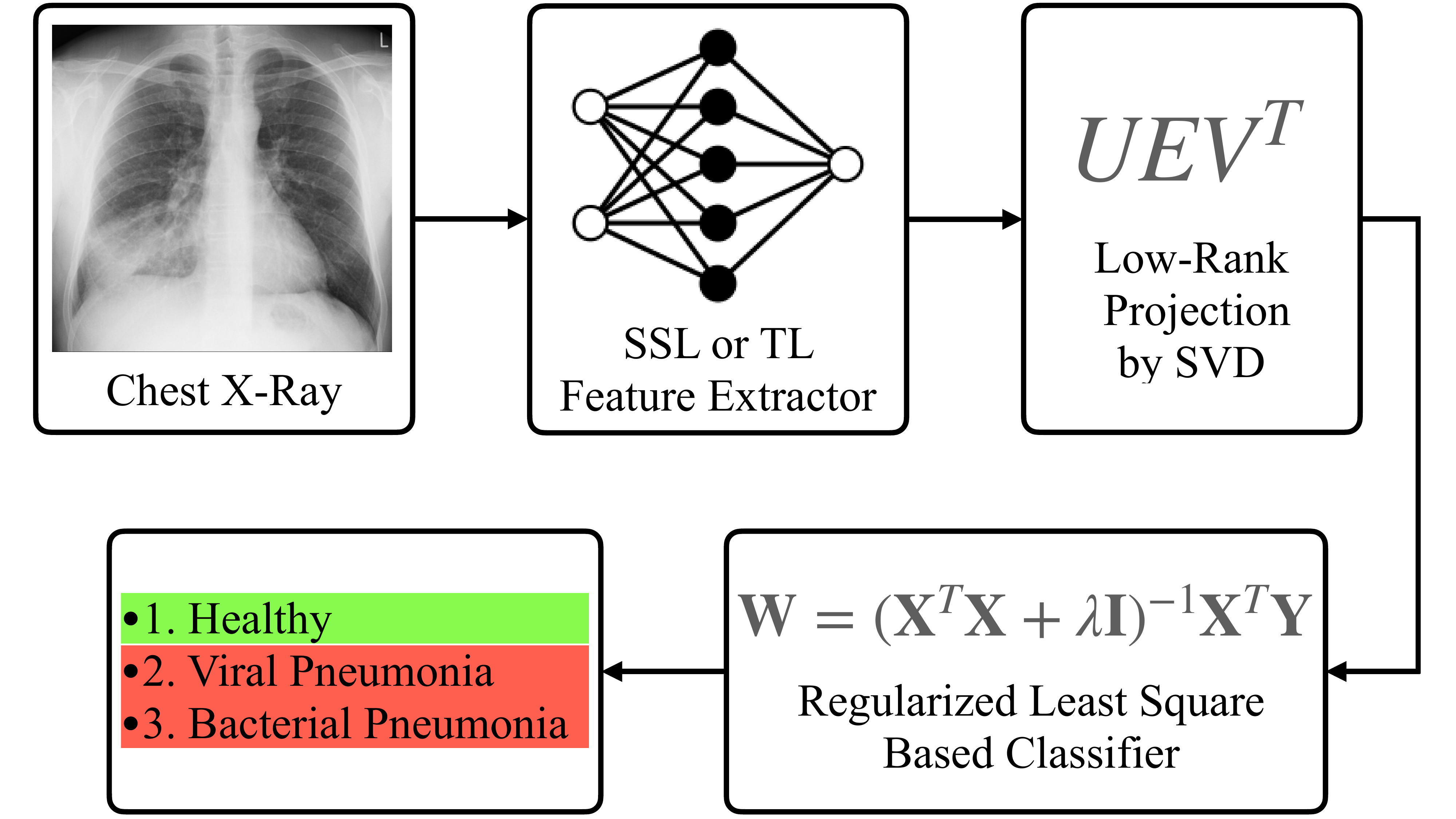}
    \caption{SVD-based least squares classification pipeline for chest X-ray pneumonia classification. The process involves extracting features using self-supervised (SSL) or transfer learning (TL) models, applying Singular Value Decomposition (SVD) for dimensionality reduction, and performing closed-form regularized least squares to classify into categories: healthy, viral, or bacterial pneumonia.}
    \label{fig:svdls_description}
\end{figure}

In this paper, we propose a method that combines the representational power of modern feature learning with the interpretability and efficiency of classical machine learning. Specifically, we introduce a Singular Value Decomposition-based Least Squares (SVD-LS) scheme for multi-class pneumonia classification. Our method leverages pretrained feature extractors from self-supervised learning (SSL) \cite{chen2020simple, jing2020self} and transfer learning \cite{tan2018efficientnet} to obtain robust image representations. These features are then fed into a closed-form SVD-based least squares classifier, enabling efficient, non-iterative training and inference (see Figure \ref{fig:svdls_description} for illustration). \\

Compared to conventional end-to-end fine-tuning, our approach offers substantial computational savings and improved numerical stability, while maintaining competitive accuracy. Moreover, explicit linear modeling of our framework enhances interpretability, an essential requirement in clinical decision support systems. We validate the efficacy of our method through comprehensive experiments on publicly available chest X-ray datasets, demonstrating its potential for scalable and reliable pneumonia classification. \\

\noindent Our key contributions are as follows:
\begin{itemize}
\setlength{\topsep}{-0.9em}
\setlength\itemsep{-0.4em} 
    \item We propose a novel application of Singular Value Decomposition-based Least Squares (SVD-LS) estimation for multi-class pneumonia classification using chest X-ray images, offering an efficient alternative to iterative deep learning classifiers.
    \item We integrate rich visual representations extracted from state-of-the-art self-supervised and transfer learning models, demonstrating that our closed-form classification approach can effectively use these features without additional gradient-based fine-tuning.
    \item We conduct comprehensive experiments comparing the performance of SVD-LS against conventional deep learning pipelines, highlighting its competitive accuracy in real-world medical imaging scenarios.
\end{itemize}

\section{Related Work}

Self-supervised learning has proven to be an effective strategy, especially in situations when labeled data is scarce or expensive to obtain. Through the utilization of self-supervised learning techniques, models can acquire valuable representations without substantial human annotation, by leveraging the inherent structure of unlabeled data. This paradigm has shown remarkable success in fields such as computer vision and natural language processing \cite{bert, he2020momentum, chen2020simple}. In particular, DINO (self-DIstillation with NO labeling) transformer models have attracted a lot of interest in the area of vision tasks \cite{dino}. DINO employs a self-distillation framework where a student network learns from a progressively updated teacher network. Though initially identical, the two networks evolve independently during training, facilitating the learning of rich and semantically meaningful visual features. Transformer-based DINO models excel at capturing both local and global structures in images, making them highly effective for self-supervised classification, segmentation, and representation learning tasks. \\

Furthermore, transfer learning has emerged as a fundamental method in machine learning, allowing pretrained models to adapt to new tasks with limited labeled data. Transfer learning uses the information gathered from large-scale datasets and challenging tasks to help models improve generalization and training efficiency when applied to new tasks. This greatly minimizes the requirement for large labeled datasets and computational resources. In particular, ResNet, introduced by He et al. \cite{resnet}, addressed the vanishing gradient problem through residual connections, enabling the training of deeper networks. Pretrained on large-scale datasets like ImageNet, ResNet has achieved strong performance in image classification and is widely used as a backbone for downstream tasks \cite{tan2018survey}. For instance, Chouhan et al. \cite{chouhan} demonstrated the effectiveness of transfer-learned CNN ensembles for pneumonia detection in medical imaging. Recent advances also combine transfer learning with self-supervised pretraining to further enhance performance in low data regimes \cite{azizi2021big}. \\

Extreme Learning Machines (ELMs) \cite{elm} are machine learning models known for their rapid training and strong generalization. They operate by randomly initializing the feature extractor, fixing input weights and biases, and then computing output weights with a least-squares solution. This method avoids iterative optimization, but relies on random projections, limiting the quality of the extracted features. Other works sought to improve efficiency by refining these random features using SVD \cite{DENG201614}; however, they still depend on random feature extractors that may lack the robustness and discriminative power of pretrained models. In our work, rather than randomly initialized feature extractors, we use either the transfer learning and self-supervised learning based models. This approach also sidesteps the significant computational demands of methods like the hybrid CNN-PCA strategy \cite{elmmedical}, which requires training the feature extractor on the target dataset before applying dimensionality reduction. \\

Our SVD-LS method combines the strengths of efficient linear classifiers with better feature representations. After extracting discriminative features from pretrained models, we apply SVD for effective dimensionality reduction and then solve a regularized least squares problem to train the classifier. This not only preserves the computational efficiency of traditional ELM frameworks, but also enhances classification accuracy, thereby providing an improved alternative for pneumonia classification.

\section{Methodology}

\subsection{SVD-Based Least Squares for Classification}

We formulate multi-class pneumonia classification as a regularized linear regression problem. Let $\mathbf{X} \in \mathbb{R}^{N \times d}$ denote the matrix of $d$-dimensional deep feature vectors for $N$ samples, and let $\mathbf{Y} \in \mathbb{R}^{N \times m}$ be the corresponding one-hot encoded label matrix for $m$ classes. The objective is to learn a linear mapping $\mathbf{W} \in \mathbb{R}^{d \times m}$ that minimizes the regularized least squares loss:
\begin{equation}
\min_{\mathbf{W}} \ \|\mathbf{X} \mathbf{W} - \mathbf{Y}\|_F^2 + \lambda \|\mathbf{W}\|_F^2,
\end{equation}
\noindent where $\lambda > 0$ is a regularization parameter and $\|\cdot\|_F$ denotes the Frobenius norm. This yields the standard ridge regression:
\begin{equation}
\mathbf{W} = (\mathbf{X}^T \mathbf{X} + \lambda \mathbf{I})^{-1} \mathbf{X}^T \mathbf{Y}.
\end{equation}

\begin{algorithm}[t]
\caption{Classification with SVD-based Least Squares}
\begin{algorithmic}[1]
\label{alg: svd-ls}
\REQUIRE Pretrained feature extractor $F$, training dataset $\mathcal{D}_\text{train}$, test dataset $\mathcal{D}_\text{test}$, regularization parameter $\lambda$, number of principal components $k$
\ENSURE Trained SVD-LS classifier $\mathbf{W}$

\STATE Extract features: $\mathbf{X}_\text{train} \gets F(\mathcal{D}_\text{train})$, $\mathbf{X}_\text{test} \gets F(\mathcal{D}_\text{test})$
\STATE Compute SVD: $\mathbf{X}_\text{train} = \mathbf{U} \mathbf{\Sigma} \mathbf{V}^T$
\STATE Reduce dimensionality: $\tilde{\mathbf{X}}_\text{train} = \mathbf{X}_\text{train} \mathbf{V}_{1:k}$
\STATE Solve closed-form regularized least squares:
\begin{equation*}
\tilde{\mathbf{W}} = (\tilde{\mathbf{X}}_\text{train}^T \tilde{\mathbf{X}}_\text{train} + \lambda \mathbf{I})^{-1} \tilde{\mathbf{X}}_\text{train}^T \mathbf{Y}_\text{train}
\end{equation*}
\vspace{-8pt}
\STATE Calculate classifier weight: $\mathbf{W} = \mathbf{V}_{1:k} \; \tilde{\mathbf{W}}$
\STATE Classify test samples: $\mathbf{\hat{Y}} ={\mathbf{X}}_\text{test} \mathbf{W}$
\RETURN Predicted labels $\mathbf{\hat{Y}}$
\end{algorithmic}
\end{algorithm}

\noindent
To reduce computational cost and mitigate overfitting, we project the features onto a lower-dimensional subspace using truncated Singular Value Decomposition (SVD). Given the decomposition $\mathbf{X} = \mathbf{U} \mathbf{\Sigma} \mathbf{V}^T$, we retain only the top-$k$ components: $\mathbf{V}_{1:k} \in \mathbb{R}^{d \times k}$ contains the top-$k$ right singular vectors, $\mathbf{U}_{1:k} \in \mathbb{R}^{N \times k}$ contains the corresponding left singular vectors, and $\mathbf{\Sigma}_{1:k} \in \mathbb{R}^{k \times k}$ is the diagonal matrix of the top-$k$ singular values. The reduced features are then given by
\begin{equation}
\label{eq: projection}
\tilde{\mathbf{X}} = \mathbf{X} \mathbf{V}_{1:k}.
\end{equation}
\noindent
We then solve the closed-form regularized least squares problem in the reduced feature space and calculate the final classifier weight $\mathbf{W}$ with a final projection:
\begin{align}
\label{eq: least_squares}
    \tilde{\mathbf{W}} &= (\tilde{\mathbf{X}}^T \tilde{\mathbf{X}} + \lambda \mathbf{I})^{-1} \tilde{\mathbf{X}}^T \mathbf{Y}, \\
    \mathbf{W} &= \mathbf{V}_{1:k} \; \tilde{\mathbf{W}}.
\end{align}
\noindent
We argue that the projection in Equation (\ref{eq: projection}) improves the numerical stability. Specifically, using $\tilde{\mathbf{X}} = \mathbf{X} \mathbf{V}_{1:k} = \mathbf{U}_{1:k} \mathbf{\Sigma}_{1:k}$, the empirical covariance of the projected features is
\begin{equation}
\mathbf{C}_{\tilde{\mathbf{X}}\tilde{\mathbf{X}}} = \frac{1}{N} \tilde{\mathbf{X}}^T \tilde{\mathbf{X}} = \frac{1}{N} \mathbf{\Sigma}_{1:k}^T \mathbf{U}_{1:k}^T \mathbf{U}_{1:k} \mathbf{\Sigma}_{1:k} = \frac{1}{N} \mathbf{\Sigma}_{1:k}^T \mathbf{\Sigma}_{1:k},
\end{equation}
which is diagonal, indicating that the features are uncorrelated. Consequently, the solution in the transformed basis becomes
\begin{equation}
\mathbf{W} = \left( \mathbf{\Sigma}_{1:k}^T \mathbf{\Sigma}_{1:k} + \lambda \mathbf{I} \right)^{-1} \mathbf{\Sigma}_{1:k}^T \mathbf{U}_{1:k}^T \mathbf{Y},
\end{equation}
\noindent
where the inversion involves a well-conditioned diagonal matrix. This provides numerical advantages even when no dimensionality reduction is applied (i.e., $k = d$). Although the condition number of the reduced system remains $\kappa(\mathbf{X}_{1:k}) = \sigma_1 / \sigma_k$, projecting onto the dominant singular vectors eliminates low-variance directions. This not only stabilizes the inversion but also improves robustness to overfitting, enabling more reliable learning in scenarios with limited or noisy training data. The use of SVD can be viewed as a form of implicit kernel mapping. Traditional kernel methods project inputs via a nonlinear function $\phi(\cdot)$ into a high-dimensional feature space that would enable linear separation:
\begin{equation}
K(\mathbf{x}_i, \mathbf{x}_j) = \phi(\mathbf{x}_i)^T \phi(\mathbf{x}_j).
\end{equation}
\noindent Instead of explicitly defining $\phi$, the SVD projects data onto an orthonormal basis defined by the directions of greatest variance. This projection captures the most discriminative structure in the data, similar to how kernel PCA \cite{scholkopf1998nonlinear, mika1998kernel} captures variance in a nonlinear space. Thus, the SVD-based projection behaves like a data-adaptive kernel, enabling more effective and computationally efficient linear classification, while maintaining robustness to feature noise. Moreover, the same least squares solution as in Equation (\ref{eq: least_squares}) can also be derived from a statistical viewpoint as the (Linear) Minimum Mean Squared Error (MMSE) estimator \cite{kay1993, kailath2000linear, erdogan2025error, erdogan2025efficient}. When the features and labels are centered, the optimal weights are given as
\begin{equation}
\mathbf{W} = \mathbf{C}_{YX} \mathbf{C}_{XX}^{-1},
\end{equation}
\noindent where $\mathbf{C}_{XX} = \mathbb{E}[\mathbf{X}^T \mathbf{X}]$ is the input covariance matrix and $\mathbf{C}_{YX} = \mathbb{E}[\mathbf{Y}^T \mathbf{X}]$ is the cross-covariance matrix between labels and inputs. This view reinforces that our approach leverages second-order statistics to learn an optimal linear predictor, offering improved interpretability and sample efficiency compared to black-box classifiers. These principles also form the basis of classical estimators like the Kalman filter \cite{kalman1960, anderson1979optimal}, which have been effectively combined with machine learning techniques; for instance, event detection tasks \cite{erdogan2024mlkf, zhang2016kalman}. \\

Our complete SVD-LS pipeline combines four key steps to achieve efficient and scalable classification with minimal training overhead. First, we extract deep features from a pretrained model; then, we apply SVD to obtain a low-rank representation. Next, we solve a closed-form regularized least squares problem to learn optimal weights, and finally, we classify new samples using these learned weights. To further clarify the procedure, Algorithm \ref{alg: svd-ls} summarizes the entire workflow. %

\subsection{Computational Complexity}

We analyze the computational cost of the proposed SVD-LS method in terms of matrix multiplication FLOPs. Given a dataset with $N$ samples, $d$ dimensional features, and $m$ output classes, the method first applies SVD to project the feature matrix $\mathbf{X} \in \mathbb{R}^{N \times d}$ onto a subspace of $k$ dimensional. Randomized SVD is used instead of exact SVD to reduce computational overhead while still accurately capturing the top-$k$ principal directions \cite{halko2011finding}. Its complexity is $\mathcal{O}\left((2q+2)\cdot Ndk + 2Nk^2 + k^2d\right)$, where the number of iterations $q$ is typically small (e.g., 3–5), which is cheaper than the $\mathcal{O}(4Nd^2)$ cost of the full SVD when $k \ll d$. Once decomposed, computing $\tilde{\mathbf{X}} = \mathbf{X} \mathbf{V}_{1:k}$ takes $\mathcal{O}(2Ndk)$, followed by computing $\tilde{\mathbf{X}}^\top \tilde{\mathbf{X}}$ at $\mathcal{O}(2Nk^2)$, $\tilde{\mathbf{X}}^\top \mathbf{Y}$ at $\mathcal{O}(2Nkm)$. Solving the regularized system costs $\mathcal{O}(k^3+2k^2m)$, followed by a final projection $\mathbf{V}_{1:k} \tilde{\mathbf{W}}$ back to the original space at cost $\mathcal{O}(2dkm)$. Overall, the total complexity of SVD-LS becomes
\begin{equation}
\mathcal{O}\left(qNdk + Nk^2 + k^2d + Nkm + k^2m + k^3 + dkm\right),
\end{equation}
with no iterative optimization required. Assuming $N \gg d \gg k \gg m$, this simplifies to $\mathcal{O}(qNdk)$. In contrast, training a linear layer with SGD or Adam optimizer involves $T = E \cdot \frac{N}{B}$ steps for $E$ epochs and batch size $B$, with each forward and backward pass costing $\mathcal{O}(2Bdm)$ FLOPs. This yields a total cost of $\mathcal{O}(T B d m)$. Since each step depends on the previous one, this process is inherently sequential and difficult to parallelize across time. In practice, when fast training is critical and the feature extractor is fixed, SVD-LS offers a substantial advantage. For instance, in our experiments in Section \ref{sec: results}, SVD-LS was $15\times$ faster than linear fine-tuning with Adam (where $E = 30$ and $B = 16$) for the DINO ViT-B/16 model. Reducing the number of epochs or increasing the batch size lowers fine-tuning time but results in a notable drop in accuracy.

\subsection{Dataset and Preprocessing}
We used the publicly available Guangzhou Women and Children’s Medical Center dataset \cite{dataa}. For preprocessing, we began by resizing the images to $224\times224$ pixels to ensure uniform input dimensions. Then, we applied mean and standard deviation normalization and performed adaptive histogram equalization (CLAHE) to enhance the model performance and reliability. In later steps, we used the training portion of the dataset to perform k-fold cross validation where we apply data augmentation to the training set of each fold for balancing the class distributions, including random horizontal flip, random rotation (10 degrees), and Gaussian blur. %

\subsection{Baselines with Pretraining and Fine Tuning Using Gradient Descent}
By using gradient descent, we trained the following models on our pneumonia classification dataset: 1) ResNet-18 trained from scratch, 2) ResNet-18 where all parameters trained after transfer learning initialization, 3) ResNet-18 where the last linear layer is fine-tuned after transfer learning initialization and 4) DINO ViT-B/16 model fine-tuned using a linear classifier. The results of these baseline models are presented in Table \ref{tab:results}.

\begin{table*}[ht]
\centering
\begin{tabular}{lcccccccc}
\toprule
 & \multicolumn{4}{c}{Train} & \multicolumn{4}{c}{Test} \\
\cmidrule(lr){2-5} \cmidrule(lr){6-9}
Model & F1-Score & Accuracy & Recall & Precision & F1-Score & Accuracy & Recall & Precision \\
 & (\%) & (\%) & (\%) & (\%) & (\%) & (\%) & (\%) & (\%) \\
\midrule
\textbf{M1: ResNet-18 Rand.} & \underline{96.17} & \underline{97.51} & \underline{95.99} & \underline{96.36} & 69.74 & 81.20 & 71.20 & 74.71 \\
\textbf{M2: ResNet-18 TL} & \textbf{97.34} & \textbf{98.22} & \textbf{97.31} & \textbf{97.38} & 73.54 & 83.76 & 74.65 & 77.56 \\
\textbf{M3: ResNet-18 FT} & 78.63 & 86.12 & 79.13 & 78.22 & \underline{77.82} & \underline{86.00} & \underline{79.35} & \underline{79.37} \\
\rowcolor{grey!20} \textbf{M4: ResNet-18 SVD-LS} & 83.19 & 89.12 & 82.57 & 84.05 & 74.69 & 84.29 & 76.53 & 78.68 \\
\rowcolor{grey!20} \textbf{M5: ResNet-50 SVD-LS} & 66.85 & 80.99 & 67.56 & 70.68 & 65.69 & 79.06 & 65.23 & 70.70 \\
\midrule
\textbf{M6: DINO FT} & 77.99 & 86.45 & 77.87 & 79.10 & 71.37 & 81.84 & 71.59 & 74.09 \\
\rowcolor{grey!20} \textbf{M7: DINO SVD-LS} & 83.59 & 89.61 & 83.40 & 84.22 & \textbf{84.27} & \textbf{90.28} & \textbf{84.49} & \textbf{85.49} \\
\bottomrule
\end{tabular}
\caption{ Average performance metrics for our models on the train and test datasets in a three-class pneumonia classification task. We evaluated several approaches: (M1) ResNet-18 trained from scratch with random initialization; (M2) ResNet-18 fully trained using transfer learning; (M3) ResNet18 with fine-tuning of the final layer via gradient descent; (M4) ResNet-18 combined with our SVD-LS method; (M5) ResNet-50 with SVD-LS; (M6) a DINO ViT-B/16-based linear classifier fine-tuned with gradient descent; and (M7) DINO ViT-B/16 with SVD-LS. Best-performing metrics are given in bold, and the second-best as underlined.}
\label{tab:results}
\end{table*}

\begin{figure}[t!]
\centering
    \subfloat[ResNet-18 TL]{\includegraphics[width=0.38\textwidth]{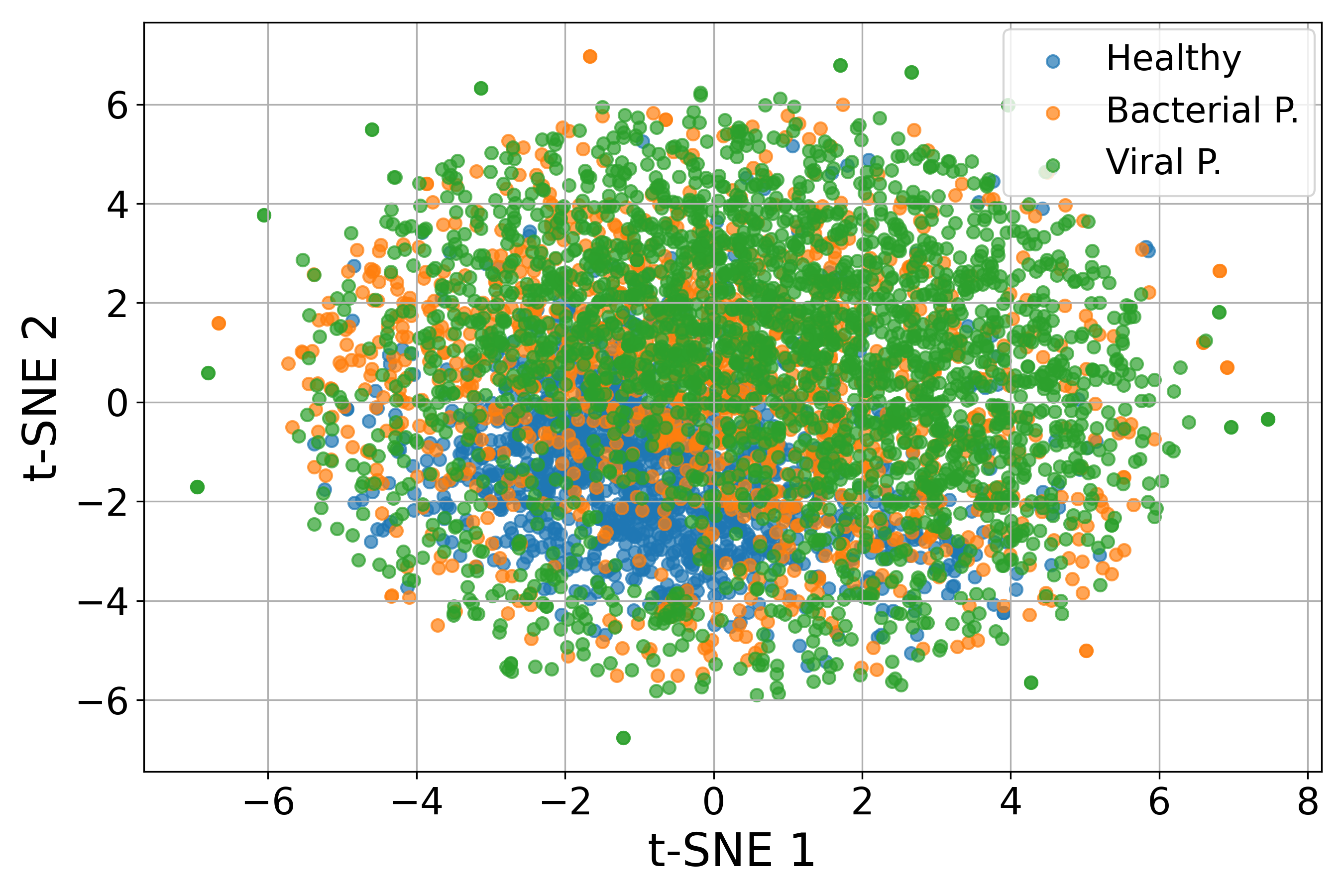}
    \label{fig:cca_mistral}} \\
    \vspace{-0.2cm}
    \subfloat[DINO ViT-B/16]{\includegraphics[width=0.38\textwidth]{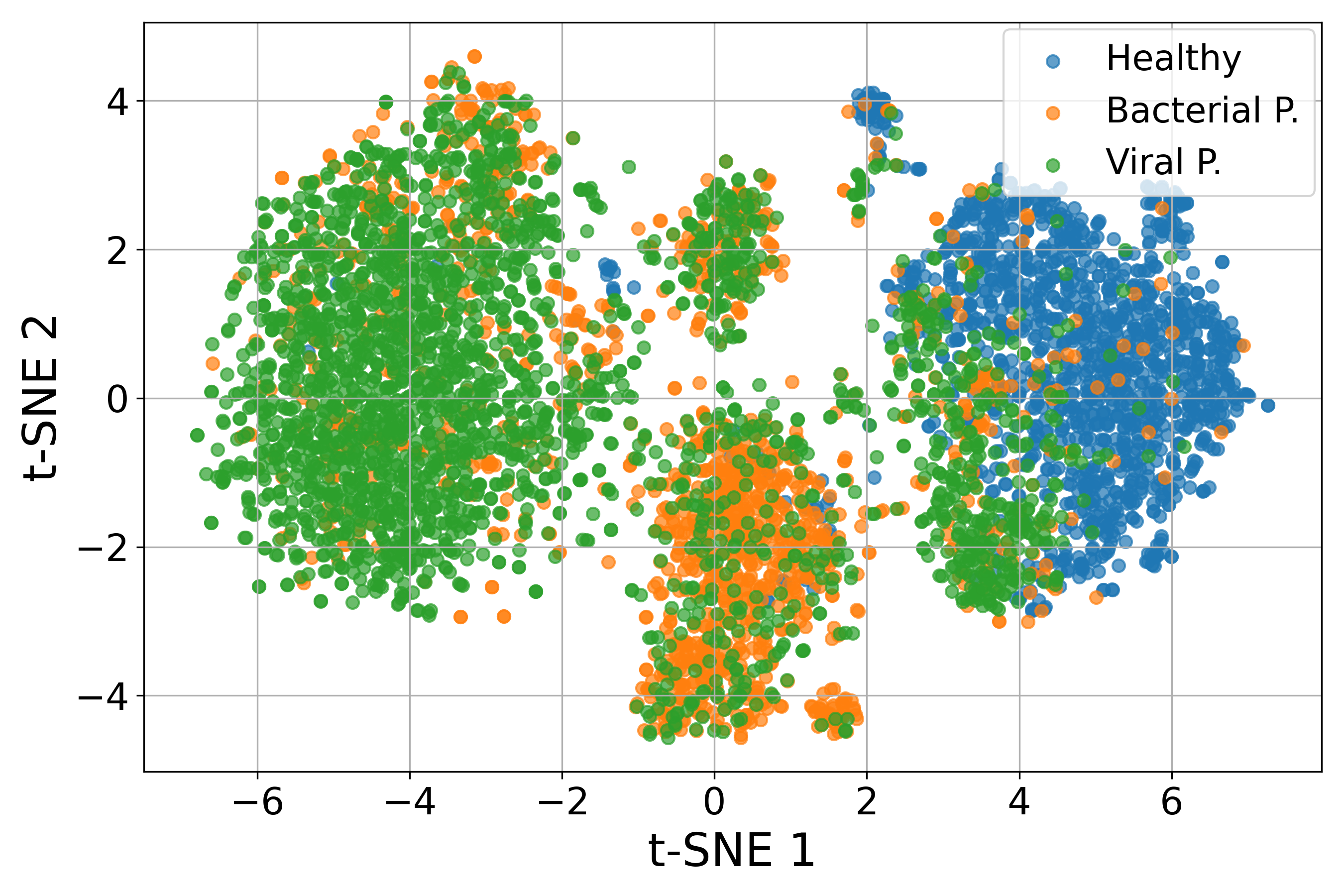}
    \label{fig:cca_llama}}
    \caption{t-SNE visualization of feature vectors extracted from (a) ResNet-18 with transfer learning and (b) DINO ViT-B/16. Each point represents a sample, colored by class label.}
    \label{fig:tsne}
\end{figure}

\subsection{Training Specifications}
In the training of each of the gradient descent-based methods, we used the Adam \cite{adam} optimizer with weight decay. We performed a grid search to find the optimal values for the hyperparameters, and k-fold cross validation with $k=5$ for each grid-search combination. The training parameters we used can be found in our GitHub repository. As we have a fixed test set, for each k-fold we split the training set into 80\% (training) and 20\% (validation) portions and reported the test set results of the model that achieved the best validation F1-score. Furthermore, for the SVD-LS classification framework, we obtained a linear classifier on top of the following models: 1) ResNet-18, 2) ResNet-50, and 3) DINO ViT-B/16. Extracted features from these models were used as inputs to our SVD-LS classifier. This allowed us to leverage the representation power of deep networks while retaining the efficiency through closed-form solutions. We also performed grid search and k-fold cross-validation with $k=5$ to determine the optimal number of principal components to retain in SVD-LS.

\vspace{-3pt}
\section{Results}
\label{sec: results}
The results of our experiments, shown in Table \ref{tab:results}, demonstrate clear differences in performance across various models for multi-class pneumonia classification on chest X-ray images. Among the evaluated models, the DINO SVD-LS model, which combines a self-supervised transformer with SVD-based least squares, outperformed all other models on the test set, achieving an F1-score of 84.27\% and an accuracy of 90.28\%. This highlights the superior generalization capabilities of self-supervised models using SVD-LS. ResNet-18 fine-tuning (M3), which involves training only the last layer, demonstrated the second-best test performance with an F1-score of 77.82\% and an accuracy of 86.00\%. While ResNet-18 SVD-LS (M4) and ResNet-50 SVD-LS (M5) used SVD-LS as well, their performance was inferior to DINO SVD-LS, suggesting that increased model complexity without corresponding feature quality may degrade performance. \\

To better understand the quality of learned features, we visualized the t-SNE \cite{tsne} projections of feature vectors from both the DINO ViT-B/16 and ResNet-18 transfer-learned models (see Figure~\ref{fig:tsne}). While both show some class separation, DINO features form more compact and well-separated clusters, indicating stronger class-discriminative structure. This makes DINO features particularly suitable for linear classifiers.  Despite their separability, these high-dimensional features may still contain redundant or noisy components. SVD addresses this by projecting them onto a lower-dimensional subspace spanned by the most informative directions, reducing overfitting and improving numerical stability. The t-SNE plots qualitatively suggest a low-dimensional manifold structure, which SVD can capture effectively, enabling the least squares classifier to generalize better in this reduced space. \\

To assess performance differences, we used the McNemar-Bowker test \cite{bowker1948test}, which compares paired classifier predictions via a chi-squared test on a contingency table. As shown in Table~\ref{tab:mcnemar}, the low $p$-values (\textless 0.05) indicate statistically significant differences between model outputs.

\begin{table}[!t]
\centering
\resizebox{\columnwidth}{!}{
\begin{tabular}{l|ccccccc}
\toprule
\textbf{Model} & \textbf{M1} & \textbf{M2} & \textbf{M3} & \textbf{M4} & \textbf{M5} & \textbf{M6} & \textbf{M7} \\
\midrule
\textbf{M1} & - & 0.000 & 0.013 & 0.000 & 0.000 & 0.000 & 0.000 \\
\textbf{M2} & 0.000 & - & 0.082 & 0.000 & 0.000 & 0.000 & 0.000 \\
\textbf{M3} & 0.013 & 0.082 & - & 0.000 & 0.005 & 0.000 & 0.001 \\
\textbf{M4} & 0.000 & 0.000 & 0.000 & - & 0.117 & 0.000 & 0.074 \\
\textbf{M5} & 0.000 & 0.000 & 0.005 & 0.117 & - & 0.000 & 0.001 \\
\textbf{M6} & 0.000 & 0.000 & 0.000 & 0.000 & 0.000 & - & 0.000 \\
\textbf{M7} & 0.000 & 0.000 & 0.001 & 0.074 & 0.001 & 0.000 & - \\
\bottomrule
\end{tabular}
}
\caption{Pairwise McNemar-Bowker test \cite{bowker1948test} results comparing the predictions of the models evaluated in Table~\ref{tab:results}.}
\label{tab:mcnemar}
\end{table}

\section{Conclusion}

In this work, we introduced an alternative classification framework for multi-class pneumonia classification using chest X-ray images, based on Singular Value Decomposition-based Least Squares (SVD-LS). Using feature representations from state-of-the-art self-supervised and transfer learning models, our approach achieves a compelling balance between accuracy and efficiency. In particular, our results demonstrated that SVD-LS combined with DINO ViT-B/16 self-supervised features yields the highest test performance among all models and baselines evaluated. Unlike conventional deep learning pipelines that rely on iterative optimization, our method provides a closed-form solution with significantly reduced computational cost and training time, making it particularly suitable for real-time applications and deployment in resource-constrained clinical environments. In future work, we plan to explore the integration of uncertainty estimation techniques into the SVD-LS pipeline to enhance its reliability in clinical decision support.

\bibliographystyle{IEEEbib}
\bibliography{refs}

\end{document}